\documentclass{article}

\usepackage{microtype}
\usepackage{graphicx}
\usepackage{subfigure}
\usepackage{booktabs} 
\usepackage{subcaption}
\usepackage{graphicx}
\usepackage{caption}
\usepackage{hyperref}
\usepackage{adjustbox}

\usepackage{todonotes}

\usepackage[accepted]{icml2024}

\usepackage{amsmath}
\usepackage{amssymb}
\usepackage{mathtools}
\usepackage{amsthm}

\usepackage[capitalize,noabbrev]{cleveref}

\theoremstyle{plain}

\theoremstyle{definition}

\theoremstyle{remark}

\usepackage{float}
\usepackage{booktabs} 
\usepackage{siunitx}

\icmltitlerunning{E(n) Equivariant Message Passing Cellular Networks}

\begin{document}

\twocolumn[
\icmltitle{E(n) Equivariant Message Passing Cellular Networks}


\begin{icmlauthorlist}
\icmlauthor{Veljko Kovac}{uva}
\icmlauthor{Erik J. Bekkers}{uva,amlab}
\icmlauthor{Pietro Liò}{comp}
\icmlauthor{Floor Eijkelboom}{uva,amlab,bo}
\end{icmlauthorlist}

\icmlaffiliation{uva}{University of Amsterdam}
\icmlaffiliation{amlab}{AMLab}
\icmlaffiliation{comp}{Department of Computer Science and Technology, University of Cambridge}
\icmlaffiliation{bo}{UvA-Bosch Delta Lab}

\icmlcorrespondingauthor{Veljko Kovac}{veljko.kovac@student.uva.nl}

\icmlkeywords{Graph Neural Networks, Topological Deep Learning, Geometric Deep Learning}

\vskip 0.3in
]
\printAffiliationsAndNotice{}

\begin{abstract}

This paper introduces $\mathrm{E}(n)$ Equivariant Message Passing Cellular Networks (EMPCNs), an extension of $\mathrm{E}(n)$ Equivariant Graph Neural Networks to CW-complexes. Our approach addresses two  aspects of geometric message passing networks: 1) enhancing their expressiveness by incorporating arbitrary cells, and 2) achieving this in a computationally efficient way with a \textit{decoupled EMPCNs} technique. We demonstrate that EMPCNs achieve close to state-of-the-art performance on multiple tasks \textit{without} the need for steerability, including many-body predictions and motion capture. Moreover, ablation studies confirm that \textit{decoupled EMPCNs} exhibit stronger generalization capabilities than their non-topologically informed counterparts. These findings show that EMPCNs can be used as a scalable and expressive framework for higher-order message passing in geometric and topological graphs.
\end{abstract}

\section{Introduction}
\label{sec:intro1}

Graph Neural Networks (GNNs) are a type of neural network designed to operate on graph-structured data, capturing relationships and interconnections between data points. They are particularly useful in domains where data is naturally represented as relations, such as social networks, molecular structures, and recommendation systems. The most common type of GNNs is a Message Passing Neural Network (MPNN), in which the representation of
nodes are updated by aggregating information in their 1-hop neighborhoods. While GNNs have shown notable success in applications such as graph classification \cite{kipf2017semisupervised}, link prediction \cite{zhang2018link}, and algorithmic reasoning \cite{Veličković2021}, their expressiveness is inherently limited by their design to capture only pairwise interactions. Some examples where modeling multi-body relationships is crucial include protein-protein interaction networks \cite{bio-ref, bio-ref2}, which require the identification of protein complexes through group interactions, and social networks \cite{social}, where the dynamics of group behavior provide essential insights. 

This limitation of GNNs is formally discussed by \citet{xu2019powerful} and \citet{morris2019weisfeiler}, where they showed that the local neighborhood aggregation used by GNNs is at most as powerful as the Weisfeiler-Lehman (WL) test in distinguishing non-isomorphic graphs \cite{weisfeiler1968reduction}.
Several studies have underscored the importance of distinguishing graphs that are indistinguishable by the 1-WL test, pointing out that, for example in molecular structures, distinct local structures around atoms should correspond to distinct representations \cite{maron-powerful, chen-wl, chai-benchmark}. One approach to overcome this problem of recognizing higher-dimensional graph features like cliques or cycles, is by generalizing MPNNs to more elaborate topological space, i.e. to take into account these features explicitly. In \citet{bodnar21a}, MPNNs are extended to operate on the clique complex of the graph -- i.e. assigning simplices to each clique in the graph -- resulting in an architecture that is provably more powerful than the WL test by creating more complex neighborhood relationships. However, as simplicial complexes have a rigid combinatorial structure that restricts the range of lifting transformations, this work was later extended to CW-complexes -- assigning features to arbitrary cells in the graph -- defining a strictly more expressive MPNN again   \cite{bodnar22}, provably not less powerful than the 3-WL test.

A generalized version of the Weisfeiler-Lehman framework has been proposed for geometric graphs, where the nodes correspond to points in Euclidean space. This framework accounts for a stronger notion of geometric isomorphism, considering spatial symmetries \cite{expressive-geom}. This so-called Geometric Weisfeiler-Leman (GWL) is used to qualify the expressivity of MPNNs with augmented discriminative capabilities of GNNs beyond the 1-WL test by integrating geometric information of the data, e.g. as done by \citet{satorras2021n} and \citet{brandstetter2021geometric}. For instance, \citet{satorras2021n} proposed $\mathrm{E}(n)$ Equivariant Graph Neural Networks (EGNN), in which the messages passed in the MPNN are conditioned on the distance between the nodes, making the model equivariant to rotations, reflections, and translations. This kind of approaches have shown to be highly effective in many domains \cite{maurice, gemnet, bekkers2024fast}.


Recently, multiple approaches to combining geometric \textit{and} topological information have been proposed \cite{eijkelboom2023mathrmen, liu2024clifford, equi-meshes}. In \citet{eijkelboom2023mathrmen}, $\mathrm{E}(n)$ Equivariant Message Passing Simplicial Networks (EMPSNs) are proposed, a method that combines simplicial message passing with geometric attributes invariant information such as angles, volume, and area, which cannot be captured when focusing only on pairwise interactions. Although this method achieved state-of-the-art results, it presents two main drawbacks. Firstly, even though the use of simplicial complexes adds higher-order topological information to the model, these structures are not the most elaborate forms available. More sophisticated structures such as CW complexes can offer a more nuanced representation of topological features, which could be especially beneficial in fields like molecular chemistry where specific topological configurations are crucial. This is specifically the case in $\mathbb{R}^3$, because due to the construction of EMPSNs, the highest-dimensional object will be a tetrahedron, i.e., a 4-body interaction. Secondly, the approach involves increased computational complexity due to the higher-dimensional message passing required in each layer. The authors use Vietoris-Rips complexes -- a higher-dimensional analogue to a radius graph -- and as such introduce a typically infeasible amount of extra features due to the combinatorial nature of such complexes, making this approach hard to use for processing large graphs or datasets where real-time performance is crucial.

In this work, we propose $\mathrm{E}(n)$ Equivariant Message Passing Cellular Networks (EMPCNs), a generalization of $\mathrm{E}(n)$ Equivariant Message Passing Simplicial Networks to cellular complexes. Our contributions are the following:
\begin{itemize}
    \item Targetting the limited expressivity of EMPSNs, EMPCNs operate on CW-complexes, hence enabling the identification of a more general set of geometric invariants across a broader range of topological objects, making them strictly more expressive than their simplicial counterparts.
   \item Addressing the increased computational costs of higher-order message passing, we introduce an efficient method to integrate higher-order topological information in geometric datasets inspired by EMPCNs. These \textit{decoupled EMPCNs} provide a simple, cellular add-on to existing MPNNs without increasing computational complexity.
   \item We show that EMPCNs achieve close to state-of-the-art performance on two tasks, outperforming many elaborate, steerable methods. Moreover, we demonstrate that the \textit{decoupled EMPCN} model provides improved performance without additional computational costs and leads to stronger generalization, providing a step towards scalable higher-order message passing.
\end{itemize}



\section{Background}
\label{sec:background}

\paragraph{Symmetry-Informed Deep Learning} Considering the intrinsic structure of data -- specifically in terms of symmetries -- is a key design principle in Geometric Deep Learning \citep{bronstein-gdl}, which has led to many successes in recent years \cite{marinka, gligorijevic2021structure}. Symmetries are mathematically described by groups, and a function respects such a symmetry -- formally: is equivariant with respect to the relevant group -- if it commutes with its group action, i.e. if
\begin{equation}
	f(g \cdot x) = g \cdot f(x) \text{ for all } g \in G.
\end{equation}
Intuitively, this means that first applying a transformation $g \in G$ and then evaluating the function is equivalent to first applying the function and then performing the transformation. A function that produces the same result when applied to both transformed and untransformed elements is referred to as group invariant.
In the setting of graph prediction, many different forms of symmetry come into play, e.g. permutations of the graphs, rotations of the graph in space (if it is embedded in some geometric space), or topological features such as rings. 

\paragraph{Message Passing Neural Networks} Message Passing Neural Networks (MPNNs) have emerged as a popular class of models for graph representation learning \cite{mpgnns}. MPNNs update node representations through the their respective 1-hop neighborhoods, i.e.
\begin{eqnarray}
h_i^{\ell + 1} = \mathsf{MPNN}(h^{\ell}_i, \{\{h^{\ell}_j\}\}_{j \sim i}),
\end{eqnarray}
where $\{\{h^{\ell}_j\}\}_{j \sim i}$ denotes the multi-set of neighbours of node $i$, and $\ell$ denotes the layer. An MPNN-layer can be decomposed into two main steps:
\begin{enumerate}
	\item  First, a message function $\mathsf{Mes}(h_i, h_j) =: m_{ij}$ computes the messages from nodes $j$ to nodes $i$, which are then aggregated in a permutation invariant manner to find the aggregated message to $i$, i.e. $m_i := \underset{j  \sim i}{\mathsf{Agg}} ~m_{ij}$.
	\item An update function $\mathsf{Update}(h_i, m_i) =: h'_i$ which updates the representation $h_i$ based on the aggregated message.
\end{enumerate}

As mentioned in Section \ref{sec:intro1}, there are two major ways to augment MPNNs: (1) by using the underlying geometry of the space in which the graph is embedded (e.g., the coordinates of atoms in a molecule), and (2) by using topological or structural information of the graph. We refer to these approaches as Geometric MPNNs and Topological MPNNs, respectively.


%

\paragraph{Geometric MPNNs} If the relevant graph is embedded in some geometric space, such as when each node is associated with coordinates \( x_i \in \mathbb{R}^n \), one can leverage the underlying geometry in the predictions, as discussed by \citet{duval2024hitchhikers}. Importantly, it is now possible to utilize more symmetries since rotating, translating, or flipping the graph leads to predictable changes. Arguably, the simplest approach to do this is $\mathrm{E}(n)$ Equivariant Graph Neural Networks (EGNNs), where the geometric information is added in the message function by conditioning it on the distance between two nodes, i.e. $\mathrm{E}(n)$ \textit{invariant} information:
\begin{equation}
	\label{eq:message_function}
	m_{ij} = \mathsf{Mes}(h_i, h_j, ||x_i - x_j||_2^2).
\end{equation}
Note that indeed distance is preserved by applying any symmetry of the Euclidean group.

%
%


\paragraph{Topological MPNNs}  Topological MPNNs  mitigate the issue of MPNNs not being able to learn higher-order structures directly by learning features on a more elaborate topological space. Two such topological spaces are simplicial and cellular complexes. 

A simplex is the generalization of a triangle to arbitary dimension, i.e. a $k$-dimensional simplex is defined by a set of $k+1$ fully connected points. As such, a graph consists of $0$-dimensional simplices (nodes) and $1$-dimensional simplices (edges), but can be trivially lifted by assigning a $k$-dimensional simplex of each clique of $k+1$ nodes. Though useful, in many domains graphs are not best understood in terms of simplices, e.g. in molecular prediction tasks where triangles are much less common than rings. A generalization of simplicial complexes is given by CW complexes, which considers arbitrary cells in the graphs, hence strictly generalizing simplicial complexes. 

More precisely, a CW complex is built from \textit{cells} of various dimensions. A cell is a space homeomorphic to an open \(n\)-dimensional disk, which informally one can understand as any object obtainable from continuously transforming an $n$-sphere. The construction of the complex involves a hierarchical glueing process:
\begin{enumerate}
    \item Start with a set of vertices (0-cells).
    \item Attach edges (1-cells) by gluing endpoints of line segments to these vertices.
    \item Attach higher-dimensional cells by mapping the boundary of each \(n\)-cell, \(S^{n-1}\), to the \((n-1)\)-dimensional skeleton \(X^{(n-1)}\) via a continuous map \(\phi: S^{n-1} \rightarrow X^{(n-1)}\).
    
\end{enumerate}

\begin{figure}[h]  

    \centering
    \includegraphics[scale=0.35]{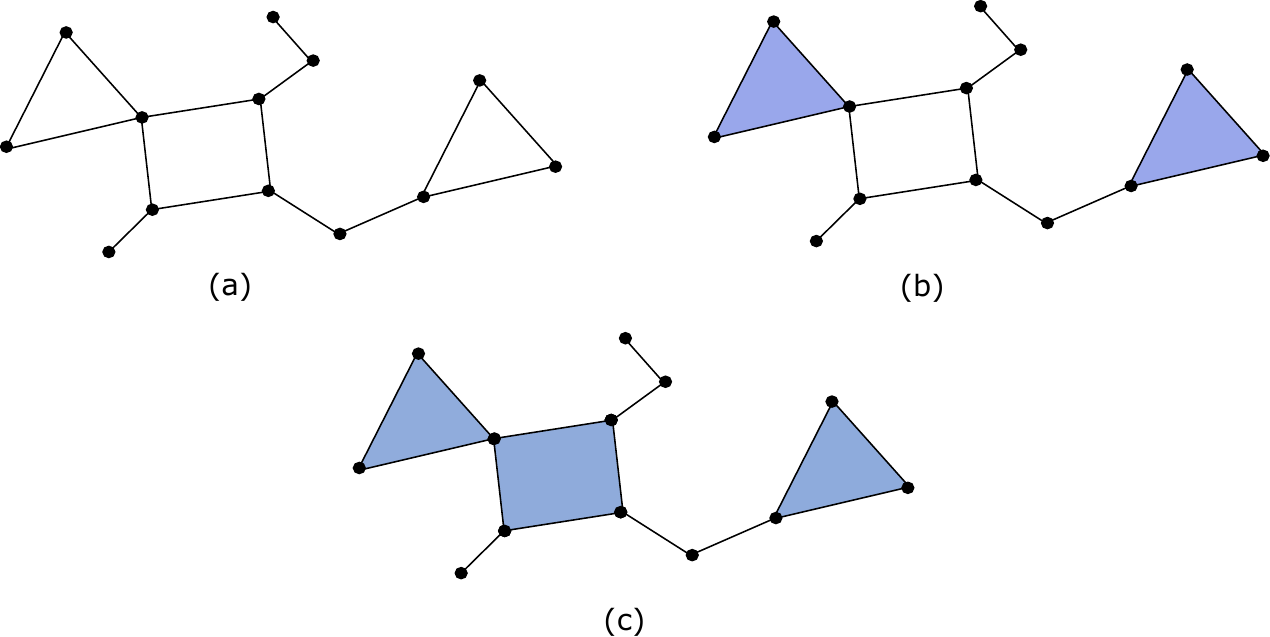}
        \caption{Visualization of a graph \textit{(a)}, a simplicial complex \textit{(b)}, and a cellular complex \textit{(c)}. As observed, a simplicial complex cannot represent arbitrary polygons.}
        \label{fig:diff-lifting}
\end{figure}

This allows for the inclusion of complex structures, such as rings, in molecular graphs as it can be seen in Figure \ref{fig:diff-lifting}.

To learn features on these topological spaces, a more elaborate set of messages are sent as introduced by \citet{bodnar22, bodnar21a}. Let $\sigma \prec \tau$ denote that $\sigma$ is a boundary of $\tau$, such as a node of an edge or an edge of a ring. With this notation, one can differentiate four types of relationships: 
\begin{enumerate}
	\item $\mathcal{B}(\sigma) = \{\tau \mid \tau  \prec \sigma \}$, i.e. all boundaries of $\sigma$, called the \textit{boundaries}.
	\item $\mathcal{C}(\sigma) = \{\tau \mid  \sigma \prec \tau\}$, i.e. all cells $\sigma$ is a boundary of, called the \textit{co-boundaries}.
	\item $\mathcal{N}_{\downarrow}(\sigma) = \{\tau \mid \exists \delta \text{ s.t. }  \delta \prec \sigma \text{ and } \delta \prec \tau \}$, i.e. the set of cells that share a boundary $\sigma$, called the \textit{lower adjacencies}. 
	\item $\mathcal{N}_{\uparrow}(\sigma) = \{\tau \mid \exists \delta \text{ s.t. }   \sigma  \prec \delta \text{ and } \tau \prec \delta \}$, i.e. the set of cells that share a co-boundary $\sigma$, called the \textit{upper adjacencies}. 
\end{enumerate}

%
%

%
%
%
The complex adjacencies defined above equip CW networks with a message passing scheme encompassing such higher-order topological features, such as vertices (0-cells), edges (1-cells), and cycles (2-cells). Consequently, updates incorporate not merely a single message, but four distinct types. However, \citet{bodnar22} demonstrated that retaining only boundary and upper adjacencies achieves equivalent expressivity, and hence the typical messages considered are:
\begin{align}
m_{\mathcal{B}}(\sigma) &= \underset{\tau \in \mathcal{B}(\sigma)}{\mathsf{Agg}}  \mathsf{Mes}_{\mathcal{B}} (h_{\sigma}, h_{\tau})\\
m_{\uparrow}(\sigma) &=   \underset{\substack{\tau \in \mathcal{N}_{\uparrow}(\sigma),\\ \delta \in \mathcal{C}(\sigma, \tau)}}{\mathsf{Agg}}  \mathsf{Mes}_{\uparrow} (h_{\sigma}, h_{\tau}, h_{\delta}) 
\end{align}

Taking molecules as an example, the first type of communication outlines how atoms communicate with bonds, and how bonds communicate with rings, whereas the second type describes interactions between atoms connected by a bond, and bonds that form a ring. 

Similar to the standard message passing framework, the update step considers the incoming messages and updates the features through a more elaborate function:
\begin{equation}
	h_{\sigma}' = \mathsf{Update}(h_\sigma, m_{\mathcal{B}}(\sigma), m_{\uparrow}(\sigma)).
\end{equation}

%

\paragraph{Geometric \& Topological MPNNs} Recently, \citet{eijkelboom2023mathrmen} combined geometric and topological GNNs by realizing that there is a direct relationship between the two: a geometric space allows one to define higher-order topological objects geometrically -- e.g. through a Vietoris-Rips complex --, while higher-order topological objects enable multi-body geometry in MPNNs. They propose $\mathrm{E}(n)$ Equivariant Message Passing Simplicial Networks, essentially combining EGNNs with simplicial message passing by defining geometric features on simplicial complexes, specifically volumes, angles, and distances, i.e. $\mathrm{E}(n)$ \textit{invariant} information. This work was later extended to the steerable case by \citet{liu2024clifford}, which defines steerable simplicial message passing using clifford algebras.

 \section{Methodology} 
\label{sec:methodology}

Building upon the $\mathrm{E}(n)$ Equivariant Message Passing Simplicial Networks, we propose $\mathrm{E}(n)$ Equivariant Message Passing Cellular Networks (EMPCNs), the generalized counterpart to EMPSNs of cellular complexes. With EMPCNs, we achieve the following two things:

\begin{enumerate}
	\item  \textbf{Enhanced Expressivity:} Similar to how normal and simplicial message passing were made geometrically equivariant through conditioning on $\mathrm{E}(n)$ invariant information, we extend these invariants to the cellular case. Since CW complexes strictly generalize simplicial complexes, this defines a more expressive MPNN.
 
	\item \textbf{Improved Scalability}: Higher-order message passing networks often struggle with scalability due to the complexity of higher-order lifting, such as EMPSNs using Vietoris-Rips complexes to construct simplices. We address this by introducing a simplified version of EMPCNs, called \textit{decoupled EMPCNs}, that explicitly learn this topological information without added computational complexity.
\end{enumerate}

\subsection{\(\mathrm{E}(n)\) Equivariant Cellular Message Passing}
\label{subsec:empcn}
As our method serves as a generalization of $\mathrm{E}(n)$ Equivariant Simplicial Message Passing, our primary focus is on making the topological objects and invariants used more expressive, rather than introducing a new method of conditioning. That is, e.g. for the boundary update
\[
m_{\mathcal{B}}(\sigma) = \underset{\tau \in \mathcal{B}(\sigma)}{\mathsf{Agg}} \, \mathsf{Mes}_{\mathcal{B}} (h_{\sigma}, h_{\tau}, \text{Inv}(\sigma, \tau)),
\]
our aim is to generalize $\mathsf{Inv}(\sigma, \tau)$, denoting the geometric invariants. 

\subsubsection{Invariants}
\label{subsec:invariants}

\paragraph{Barycentric Subdivision}

\begin{figure}[H]
	\centering
	\includegraphics[scale = 0.55]{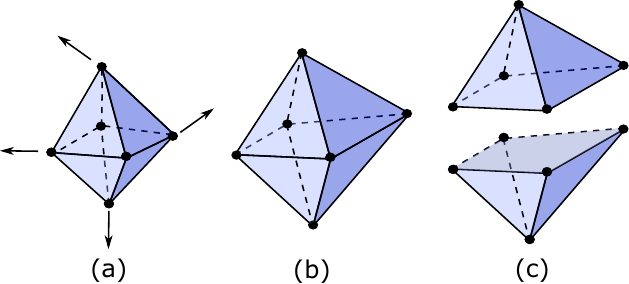}

	\caption{Change of invariants after position updates: \textit{a)} Initial graph with arrows indicating the future displacement of the respective nodes. \textit{b)} Displaced graph showing the updated positions of the nodes. \textit{c)} Cell decomposition into two simplices.}
 \label{fig:moving-volume}
\end{figure}

Crucially, we make the observation that we can understand the cells in a CW complex as a set of simplices glued together. Formally, we can define all their geometric invariants using them. To explain this, we refer to Barycentric subdivision \cite{barycentric1}, a technique typically employed to decompose a simplex into smaller simplices, which can also be applied to CW complexes. Specifically, a regular CW complex can be subdivided into a simplicial complex, resulting in a structure where the simplices are efficiently combined into closed cells, as shown in Figure \ref{fig:moving-volume}.

\vspace{-0.3pt}

This is key to our approach, as it allows us to directly generalize our existing invariants in EMPSN to the cellular case. 

\paragraph{Generalized Volume and Area}
In the context of graphs embedded within an \(n\)-dimensional Euclidean space, the convex hull of some set $S$ is defined as the smallest convex set containing all the points in $S$.  This set -- and its volume -- are typically obtained through the Quickhull algorithm \citep{quickhull}, an efficient method in computational geometry that relies on decomposing the hull into simplices, as their volumes are easy to compute, and then summing these subvolumes. For a \(n\)-simplex $\xi$ defined by vertices \(\xi = (x_0, x_1, \ldots, x_n)\), the volume is given by:
\[\text{Vol}(\xi) = \frac{1}{n!} \left| \det \begin{pmatrix}
x_1 - x_0 & \cdots & x_n - x_0
\end{pmatrix} \right|\]
As the total volume of the hull is the sum of the volumes of all the simplices, note that we trivially obtain the volume of a single simplex through this method, hence strictly generalizing EMPSNs. 

Following a similar argument, the surface area of the hull is computed by dividing its area into \((n-1)\)-simplices, and again the total surface area is obtained by summing the areas of all $(n-1)$ simplices forming the hull's surface. Note that again we trivially obtain the simplicial case from here.

\paragraph{Other Invariants} For all other types of message passing, one could use the invariants introduced by \citet{eijkelboom2023mathrmen}, which range from distances between two nodes to calculating the dihedral angles for edge-to-edge communication, as these are not affected by a change in topological space. A detailed list of invariants for each experiment can be found in Appendix \ref{app:details}.

\paragraph{Scalable Convex Hull Estimation}
Computing the convex hull and its invariants during training can be computationally intensive, especially when nodes are moving and their positions are constantly updated, requiring re-calculation of invariants in every iteration. To address this, we can approximate these invariants by assuming the radius of the ring to be half of the maximum pairwise distance between nodes in the ring. By doing so, we are essentially taking the radius of a sphere in \(\mathbb{R}^3\) that includes all the points in the ring. This simplification allows for faster invariant calculations, making it more feasible to track changes in invariants due to node movement.

\subsubsection{Geometric Weisfeiler-Lehman}

As introduced in Section \ref{sec:intro1}, the Geometric Weisfeiler-Leman framework can be used to compare the expressivity of different geometric MPNNs. The three axes of the plot are the \textit{body order}, \textit{tensor order}, and \textit{depth}, which corresponds to whether equivariant layers are being used. Using this framework as a reference, we position our proposed method, EMPCNs, at least as high as EGNNs \cite{satorras2021n} and PaiNN \cite{schutt2021equivariant}, given that our features transform equivariantly within the Cartesian system. However, our increased expressiveness comes from the axis corresponding to body order, which denotes the number of nodes needed to compute the geometric invariant information. Since our method calculates the $n$-dimensional volume, it requires $n+1$ points and thus can be placed arbitrarily high in terms of body order. Even when restricting ourselves to $\mathbb{R}^3$, our method remains more expressive than EGNN, demonstrating its ability to capture complex geometric information.

\subsection{Decoupled EMPCNs}
\label{subsec:topo-egnn}

As mentioned in Section \ref{sec:intro1}, the most straightforward approach when dealing with geometric graphs is to interconnect all nodes. This allows the network to identify relevant edges on its own, similar to a Graph Transformer architecture \cite{gps, gt-attn, dwivedi2021generalization}. Clearly, the number of cells in a graph grows combinatorially with more edges, meaning that EMPCNs cannot be applied in a feasible way.  

\begin{figure}[h]  
    \centering
    \includegraphics[scale = 0.3]{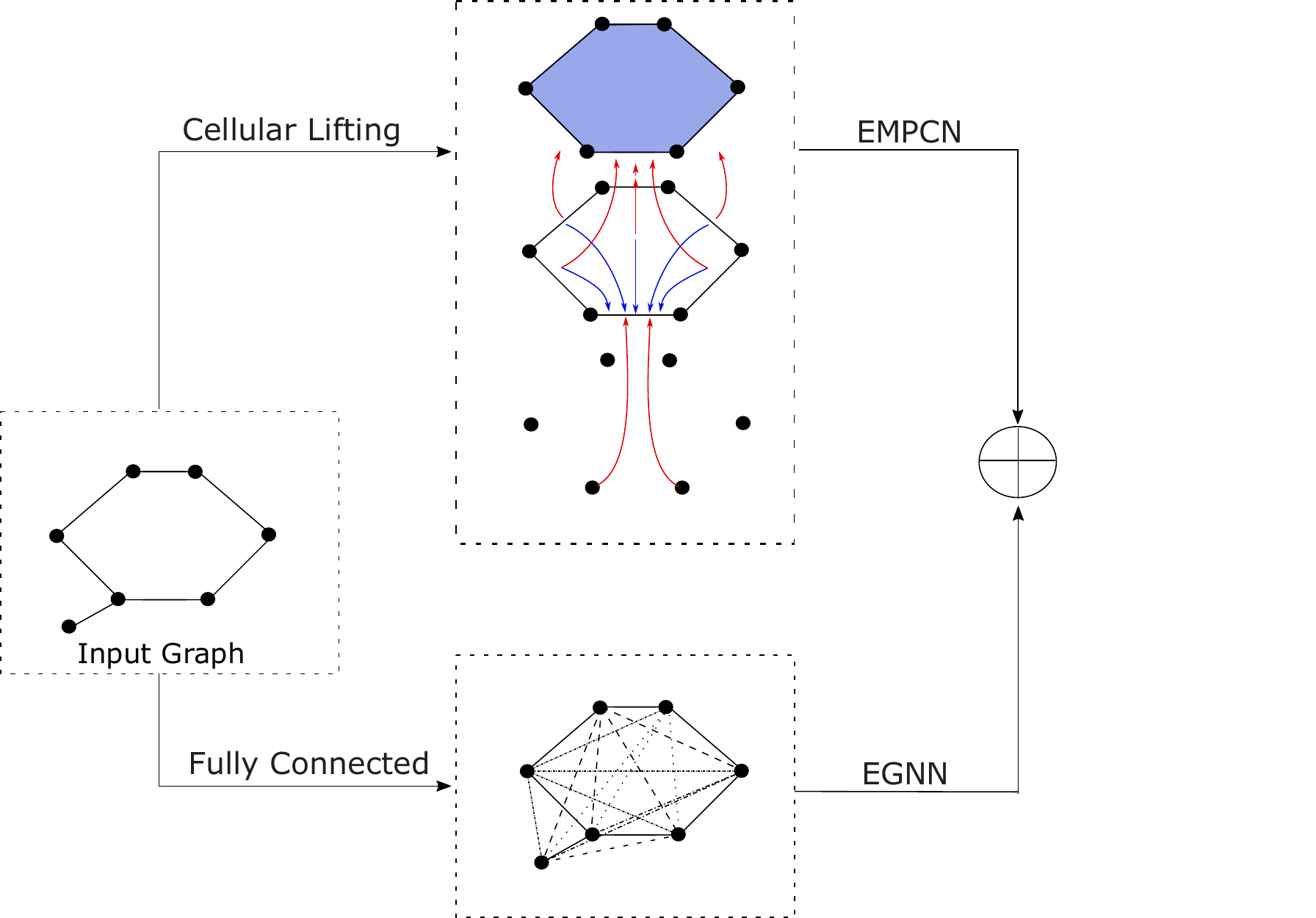}
        \caption{Pipeline of \textit{Decoupled EMPCNs}. The input graph is split into two: the cellular lifted graph for higher-order message passing and the fully connected graph for direct node communication.}
    \label{fig:decoupling}
\end{figure}

To this end, we propose a scalable cellular message passing framework called \textit{decoupled EMPCNs}, which essentialy serves as an add-on to existing fully connected geometric MPNNs. \textit{Decoupled EMPCNs} operate on 1) a node graph facilitating direct node communication and 2) a cellular lifted graph derived from the input graph, enabling higher-order message passing based on the input's graph topology. The decoupling procedure can be seen in Figure \ref{fig:decoupling}.

In a fully connected setting -- which is the typical setting for such tasks -- the number of message sent is $\mathcal{O}(|\mathcal{V}|^2)$. To obtain an equal computation complexity, we introduce a new type of cellular adjacency:
\begin{equation}
	\mathcal{P}(\sigma) := \{\tau \mid \tau \subset \sigma \text{ and } |\tau| = 1 \},
\end{equation}
i.e. all vertices that make up $\sigma$. Clearly, in the typical case of e.g. molecular prediction tasks, the number of rings $|\mathcal{R}|$ is much smaller than the number of nodes in the graph. If we therefore only perform message passing over the fully connected adjacency for the nodes and over $\mathcal{P}(\sigma)$ for the highest-order features we want to learn, the amount of message sent is:
\begin{equation}
	\mathcal{O}(|\mathcal{V}|^2 + |\mathcal{R}|\cdot|\mathcal{V}|) =  \mathcal{O}(|\mathcal{V}|^2) \text { if }  |\mathcal{R}| < |\mathcal{V}|.
\end{equation}

For example, in molecular prediction tasks, this adjacency allows the model to directly propagate information between rings and their adjacent nodes. By incorporating this additional information, which is further enhanced by geometric invariants such as volumes, the model can better understand how adjacent nodes influence ring properties and vice versa. Since functional groups  attached to rings -- e.g. hydroxyl, methyl, nitro groups for molecules -- can significantly affect ring geometry, calculating these geometric properties could provide a better insights into the complex interactions and structural variations around the rings.


Note that our proposed method is not limited to cellular lifting methods and ring structures. Instead, it allows for the definition of any type of higher-order structure as a template, followed by the calculation of its geometric invariants. This method serves as a general framework that combines geometric GNNs with additional explicit topological information, offering scalability and flexibility in including various higher-order structures.

\section{Related Works}
\label{sec:related}

\paragraph{Higher-Order Networks:} Recent works have expanded Graph Neural Networks (GNNs) to include simplicial and cellular complexes \cite{papillon2024architectures}, where especially the work by \citet{hajij2021cell} and \citet{bodnar22} are closely related to this work. To our knowledge, \citet{eijkelboom2023mathrmen} is the first work explicitly integrating topological and geometric MPNNs and as such form the basis of our approach. A similar approach was later proposed to integrate mesh face geometric information into $\mathrm{E}(3)$ Equivariant GNNs by altering the update function \cite{equi-meshes}. Furthermore, more geometrically complex architectures utilizing Clifford algebras for simplicial complexes have been introduced \cite{liu2024clifford}. However, here, the improvement in performance comes at a significantly high computational cost.

\paragraph{Efficient Higher-Order Networks:} Addressing the computational complexity inherent in higher-order message passing, recent studies have sought ways to enhance efficiency. Notably, SaNN \cite{gurugubelli2023sann} introduced a method to integrate simplicial inductive bias into networks efficiently, maintaining constant training time and memory requirements. This was achieved by enhancing neural models with pre-aggregated features of simplices across different orders. A different direction was taken by Topo-MLP \cite{ramamurthy2023topomlp}, where they learn representations for simplices across multiple dimensions through the adoption of a higher-order neighborhood contrastive loss, thereby moving away from the traditional message passing framework.

\section{Experiments \& Results}
\label{sec:res}

For all experiments, the implementation and experimental details are provided in Appendix \ref{app:implementation-details} and \ref{app:details} respectively.

\subsection{N-Body System}
\label{subsec:nbody}

As introduced by \citet{nri-kipf}, the N-body system experiment involves tracking the trajectories of five charged particles in three-dimensional space over time. The objective is to predict the positions of all particles after 1,000 time steps, starting from their initial positions and velocities. To ensure a consistent comparison, we adopted the experimental setup as described by \citet{satorras2021n}. While the system operates according to physics, with particles being attracted or repelled based on their charges,  the task remains E(n) equivariant, maintaining these symmetries throughout the entire trajectory. This means that the predictions are consistent under Euclidean transformations, such as rotations and translations, ensuring the model respects these inherent physical symmetries.

\begin{table}[h]
\centering
\caption{Mean Squared Error for the N-body system experiment.}
\begin{tabular}{lc}
\hline
Model &  \textbf{MSE ($\downarrow$)} \\
\hline
TFN \cite{thomas2018tensor} & 0.0155\\
PONITA \cite{bekkers2024fast} & 0.0043 \\
SEGNN  \cite{brandstetter2021geometric} & 0.0043 \\
CGENN\cite{ruhe2023clifford} & \textbf{0.0039}\\
EMPSN \cite{eijkelboom2023mathrmen} & 0.0063\\
EGNN \cite{satorras2021n} & 0.0071     \\
\hline
EGNN + CIN & $0.0057 \pm 0.0002$\\
EMPCN & $0.0046 \pm 0.00005$ \\

\hline
\end{tabular}

\label{table:mse-nbody}
\end{table}

\begin{table*}[tb]
\centering
\caption{QM9 MAE ($\downarrow$) Results. Relative improvement with respect to EGNNs (gain) is provided for comparison.}
\begin{adjustbox}{max width=\textwidth}
\begin{tabular}{c c c c c c c c c c c c c}
\hline
Task & $\alpha$ & $\Delta \varepsilon$ & $\varepsilon_{\text{HOMO}}$ & $\varepsilon_{\text{LUMO}}$ & $\mu$ & $C_v$ & $G$ & $H$ & $R^2$ & $U$ & $U_0$ & ZPVE \\
Units & (bohr$^3$) & (meV) & (meV) & (meV) & (D) & (cal/mol K) & (meV) & (meV) & (bohr$^3$) & (meV) & (meV) & (meV) \\
\hline
NMP & .092 & 69 & 43 & 38 & .030 & .040 & 19 & 17 & .180 & 20 & 20 & 1.50 \\
SchNet  & .235 & 63 & 41 & 34 & .033 & .033 & 14 & 14 & .073 & 19 & 14 & 1.70 \\
Cormorant & .085 & 61 & 34 & 38 & .038 & .026 & 20 & 21 & .961 & 21 & 22 & 2.02 \\
TFN & .223 & 58 & 40 & 38 & .064 & .101 & - & - & - & - & - & - \\
SE(3)-Tr. & .142 & 53 & 35 & 33 & .051 & .054 & - & - & - & - & - & - \\
DimeNet++  & \textbf{.043} & \textbf{32} & 24 & 19 & .029 & .023 & 7 & \textbf{6} & .331 & 6 & 6 & 1.21 \\
SphereNet  & .046 & \textbf{32} & \textbf{23} & \textbf{18} & .026 & \textbf{.021} & 8 & \textbf{6} & .292 & 7 & 6 & 1.21 \\
PaiNN & .045 & 45 & 27 & 20 & \textbf{.012} & .024 & 7 & \textbf{6} & .066 & \textbf{5} & \textbf{5} & \textbf{1.12} \\
SEGNN & .060 & 42 & 24 & 21 & .023 & .031 & 15 & 16 & .660 & 12 & 15 & 1.62 \\
MPSN & .266 & 153 & 89 & 77 & .101 & .122 & 31 & 32 & .887 & 33 & 33 & 3.02 \\
EMPSN & .066 & 37 & 25 & 20 & .023 & .024 & \textbf{6} & 9 & .101 & 7 & 10 & 1.37 \\
EGNN & .071 & 48 & 29 & 25 & .029 & .031 & 12 & 12 & .106 & 12 & 12 & 1.55 \\
\hline
\textbf{Decoupled EMPCN} & .063 & 40 & 27 & 22 & .026 & .027 & 10 & 10 & .104 & 10 & 11 & 1.51 \\
\textbf{Gain} & 11\% & 17\% & 7\% & 12\% & 10\% & 13\% & 17\% & 17\% & 2\% & 17\% & 9\% & 3\% \\
\hline
\end{tabular}
\end{adjustbox}
\label{tab:qm9-mae}
\end{table*}

While our method can be applied to $\mathrm{E}(n)$ equivariant scenarios, for this experimental setup, we work with $\mathrm{E}(3)$ equivariance. As seen in Table \ref{table:mse-nbody}, our performance is close to on par with state-of-the-art approaches.

To evaluate the significance of geometric invariants in higher-order message passing, we conducted an additional experiment with a modified version of Cellular Message Passing (CIN) \cite{bodnar22}. In this setup, the node-to-node communication (0-cell to 0-cell) included the distance norm, transforming it into EGNN. However, the higher-order message passing layers did not include any geometric invariants, distinguishing it from our proposed EMPCNs. As shown in Table \ref{table:mse-nbody}, while this approach outperforms EGNN alone, it falls short by 20\% in performance compared to EMPCNs, highlighting the important role of geometric invariants in the higher order message passing layers. It is worth noting that the number of parameters has been kept similar to EGNN in all of our experiments.

\subsection{QM9}
\label{subsec:qm9}
The QM9 dataset, first introduced by \citet{qm9} and subsequently studied by \citet{mpgnns} and \citet{moleculenet}, comprises approximately 130,000 graphs, each consisting of around 18 nodes. In this dataset, the graphs represent molecules, with the nodes representing atoms, and edges representing various types of bonds between these atoms. The objective of analyzing this dataset is to predict quantum chemical properties. 

For this experiment, we followed the same setting as EGNN \cite{satorras2021n}. We applied our proposed method as introduced in Section \ref{subsec:topo-egnn}, called \textit{decoupled EMPCNs}, using EGNN for the fully connected node-to-node communication, while employing EMPCNs for the higher-order message passing. As discussed in Section \ref{subsec:topo-egnn}, different adjacencies and geometric invariants can be used. For this task, we specifically chose only the ring-to-node communication and its respective invariants, including the distance from each node to the ring's midpoint, the ring's length, volume and area.

Note that, given the data consists of small molecules -- where cells are not present much --  it is crucial to optimize parameter usage to avoid unnecessary computations. We hence opt to allocate 75\% of the parameters to node-to-node and 25\% to ring-to-node message passing, such that the majority of parameters are allocated to the main architecture, which, in our case, is EGNN.

As indicated in Table \ref{tab:qm9-mae}, incorporating this additional ring information  yielded an average improvement of approximately 10\% over the baseline EGNN model. While one could potentially generate additional rings via the RadiusGraph or Vietoris-Rips complex, our experiment demonstrates that even with the initial graph’s topology, we can achieve significant performance gain \textit{without} extra computational complexity. The reason for not applying the traditional EMPCNs in this experiment is because, in the fully connected setting, the number of rings becomes extremely large, and calculating the respective invariants is computationally infeasible.

\subsection{CMU Motion Capture}
\label{subsec:cmu}

In this experiment, we evaluated our \textit{decoupled EMPCN} model using the CMU Motion Capture Database \cite{cmu}, which tracks the trajectories of human motion in various scenarios. The dataset's graphs are composed of 31 connected nodes, each corresponding to a specific position on the human body during walking. Our goal is to forecast the node positions 30 timesteps into the future based on the positions from a randomly chosen frame. We adopted a similar framework to \citet{kipf2017semisupervised, huang2022equivariant,liu2024clifford}, concentrating on the walking motion of a single subject (subject \#35) over 23 trials.

\begin{table}[H]
\caption{Comparison of different methods based on MSE ($10^{-2}$) for the CMU Motion Capture Dataset.}
\centering
\begin{tabular}{lc}
\hline
\textbf{Method} & \textbf{MSE ($\downarrow$)} \\
\hline
TFN \cite{thomas2018tensor}& 66.9 \\
SE(3)-Tr \cite{fuchs2020se3transformers}& 60.9 \\
GNN \cite{mpgnns} & 67.3 \\
GMN (200K) \cite{huang2022equivariant} & 17.7 \\
EMPSN (200K) \cite{eijkelboom2023mathrmen} & 15.1 \\
CGENN (200K) \cite{ruhe2023clifford} & 9.41 \\
CSMPN (200K) \cite{liu2024clifford} & \textbf{7.55} \\
EGNN (200K) \cite{satorras2021n} & 31.7 \\
\hline
Decoupled-EMPCN (200K)  & 9.26  $\pm$ 0.11\\
\hline
\end{tabular}

\label{tab:cmu}
\end{table}

To construct our dataset, we followed a similar approach as outlined in Section \ref{subsec:qm9}. For node-to-node communication, we employed a traditional EGNN architecture, where all nodes are interconnected. However, for cellular lifting, due to the absence of obvious rings, we adopted a methodology similar to that introduced by \citet{liu2024clifford}. We manually connected nodes in our input graph that we believe will provide additional information. Specifically, in addition to connecting elbow joint nodes with shoulder and palm nodes to create edges and triangles, and linking hip, knee, and heel nodes to form another set of edges and triangles as proposed, we introduced rings of length four. These rings connect the two elbows with the two knees, two arms with two wrists, two knees with the two feet, and two knees with the feet and the hip. While rings of varying lengths and structures could be utilized, the ultimate goal is to connect nodes that may be significant for the downstream task. Once these ring structures are constructed, cellular lifting is performed, which is then efficiently utilized by our \textit{decoupled EMPCNs} as discussed in Section \ref{subsec:topo-egnn}. Similar to the experiment described in Section \ref{subsec:qm9}, we used node-to-node and ring-to-node.

As shown in Table \ref{tab:cmu}, our proposed model achieves scores that are nearly state-of-the-art without relying on highly computationally intensive techniques such as steerable methods or Clifford algebra methods. This demonstrates that our approach, which simply adds an extra message passing layer to learn topological features of the graph combined with the corresponding geometric invariants, can significantly improve the performance of GNNs on geometric datasets.

\subsection{Ablation Study}

\begin{figure}[h]
    \centering
\includegraphics[width=0.45\textwidth]{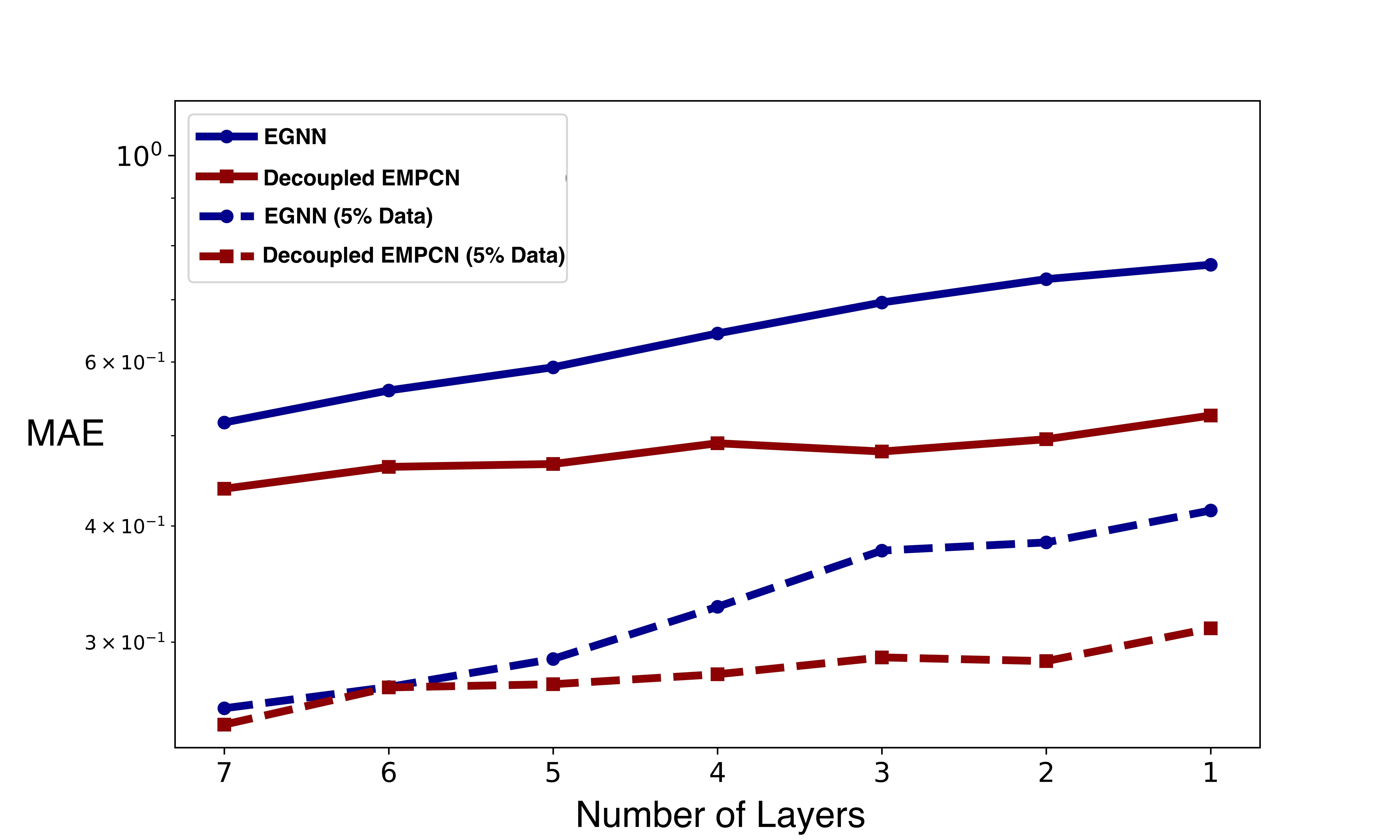}
    \caption{Ablation study on the \(\alpha\) property of QM9, comparing decoupled EMPCN (red) and EGNN (blue). Both models were trained with 100\% data (solid lines) and 5\% data (dashed lines), and evaluated on MAE (\(\downarrow\)) across different numbers of layers.}
    \label{fig:res_ab}
\end{figure}

In our ablation study, we compared the performance of EGNN and \textit{decoupled EMPCN} under varying conditions of data availability and model complexity. We evaluated both models in a small-data regime to test generalizability and in a small-model regime to assess parameter efficiency. The results are summarized in \autoref{fig:res_ab}.

First, we notice that as the number of layers decreases -- i.e. a simpler model is considered -- the EGNN baselines perform worse than their cellular counterparts, and this performance gap increases as the number of layers decreases. However, when a large enough model \textit{and} enough data are used, this gap can be almost closed. 
That being said, when given access to only $5\%$  of the data, EGNN is consistently outperformed, even in large-model settings. This indicates that the cellular counterpart is not only more data-efficient but also has significantly stronger generalization capabilities.

 These findings highlight the robustness and efficiency of decoupled EMPCNs, illustrating that they are a more reliable choice for scenarios where data and computational resources are limited as is typical in real-life settings. Robustness and efficiency here refer to maintaining higher performance than EGNN across different data regimes and model complexities.




\section{Conclusion}

In this study, we introduced $\mathrm{E}(n)$ Equivariant Message Passing Cellular Networks (EMPCNs), extending the capabilities of EMPSNs by transitioning from simplicial to cellular complexes. This transition allows for a more flexible representation of topological structures, enabling the inclusion of complex features such as rings in molecular graphs. Consequently, EMPCNs can define geometric invariants over a broader range of objects, enhancing the expressiveness of the model.

We also proposed a method to efficiently integrate topological information into geometric graphs that allow for a fully connected backbone in the standard message-passing phase. By decoupling the input graph into a fully connected graph for direct node-to-node communication and a cellular lifted graph for higher-order message passing, we combined the benefits of both approaches. This technique maintains the original graph’s topology while utilizing any MPNN architecture for direct communication, thereby improving both performance and expressivity.



In summary, decoupled EMPCNs provide a scalable and expressive framework for higher-order message passing in geometric and topological graphs. Although this method is proposed for cellular complexes, the same techniques could be adapted to other higher-order path message passing methods, potentially enhancing the expressiveness by incorporating various structures beneficial for specific tasks. We believe that exploring the intersection between geometric and topological MPNNs is a crucial aspect of developing efficient and strongly generalizing models, especially in low-data settings.

\paragraph{Limitations and Future Work} The main limitation of EMPCNs is the computational cost associated with the calculation of the introduced geometric invariants. Although it is possible to retain only those invariants relevant to a specific downstream task, an interesting direction for future research would be to pre-aggregate cell representations and compute the geometric invariants solely for key structures essential to the task. However, it is worth mentioning that these invariants are significantly cheaper to compute than the ones in methods using e.g. Clifford algebras.

For the decoupled EMPCNs, the primary limitation is that in geometric graphs, initial connectivity may not be available, and only a point cloud may be provided. While pre-domain knowledge, as discussed in Section \ref{subsec:cmu}, can be used to identify important higher-order structures, a promising research direction would be to develop methods to autonomously identify these structures based on the downstream task. In addition, as we just showed with the decoupled EMPCNs a generic template that can combine geometric and topological approaches, our goal was to beat the performance of the main architecture being used, which was EGNN in our case. Further experiments with more advanced architectures like SEGNN \cite{brandstetter2021geometric} should be tested. Finally, although our method is proposed for cellular complexes, the same technique could be applied to higher-order path message passing, potentially resulting in a more expressive framework by incorporating different types of structures that are known to be useful for a particular downstream task.

\newpage
\bibliography{empcn_update}
\bibliographystyle{icml2024}

\newpage
\appendix
\onecolumn

\section{Implementation Details}
\label{app:implementation-details}

This section describes the implementation of EMPCNs. After a cellular lifting is performed, we have : 1) a set of features for each cell 2) adjacency relationships between these cells 3) positional information of the nodes \(\{x_i\}\), and 4) optionally, initial velocities of the nodes \(\{v_i\}\). The learnable functions are as follows:

\begin{itemize}
    \item Feature Initialization: Initial features are embedded using a linear embedding process. For each dimension \(n\), we apply a separate embedder: 
  \[ \{\text{Initial Feature}_n \rightarrow \{ \text{LinearLayer}_n \} \rightarrow \text{Embedded Feature}_n\} \]
  \item Message Passing: For an adjacency \(A\) from \(\tau\) to \(\sigma\), we learn a message function. Using the input \([h_\sigma \oplus h_\tau \oplus \text{Inv}(\sigma, \tau)]\), where \(\oplus\) denotes concatenation, the message function is defined as follows: 
  \[ \{[h_\sigma \oplus h_\tau \oplus \text{Inv}(\sigma, \tau)] \rightarrow \text{LinearLayer} \rightarrow \text{Swish} \rightarrow \text{LinearLayer} \rightarrow \text{Swish} \rightarrow m_{\sigma,\tau}^A\} \] 
  \item Edge Inference: For each message, we compute the edge importance similar to \citet{satorras2021n}:
  \[ \{m_{\sigma,\tau}^A \rightarrow \text{LinearLayer} \rightarrow \text{Sigmoid}\} \rightarrow e_{\sigma,\tau}^A \]
  \item Cell Update: For each cell, an update is learned through the different adjacency communications: 
  \[ [h_\sigma, \{m_{\sigma,\tau}^A\}, \{e_{\sigma,\tau}^A\}] \rightarrow \{ h_\sigma \oplus \left( \bigoplus_{A_{\sigma, \tau}} e_{\sigma,\tau}^A \cdot m_{\sigma,\tau}^A \right) \rightarrow \text{LinearLayer} \rightarrow \text{Swish} \rightarrow \text{LinearLayer} \} \rightarrow h_\sigma' \]
  \item Readout: To get the final prediction, a pre-readout and readout phase take place:
\[ \{h_\sigma'^n\} \rightarrow \text{LinearLayer} \rightarrow \text{Swish} \rightarrow \text{LinearLayer} 
 \rightarrow \bigoplus_n \left( \sum_i h_i'^n \right) \rightarrow \text{LinearLayer} \rightarrow \text{Swish} \] 
 \[  \rightarrow \text{LinearLayer} \rightarrow \text{Prediction} \]

\end{itemize}

This architecture design is related to the $\mathrm{E}(n)$ Equivariant Cellular Message Passing Cellular Networks introduced in Section \ref{subsec:empcn}. In the \textit{decoupled EGNN} discussed in Section \ref{subsec:topo-egnn}, only 0-cells (nodes) and 2-cells (rings) are used in the experiment described in \ref{subsec:qm9}. The message passing and update rules are adapted to handle the interactions directly between these two types of cells.

In the experiments of N-Body and CMU Motion from Sections \ref{subsec:nbody} and \ref{subsec:cmu} respectively, where the final prediction is the displaced position in the next timestep, and the velocity of each node is used within the prediction, the equations for the position update are similar to the ones introduced by \citet{satorras2021n}:

\begin{itemize}
    \item Velocity Update:
      \[ v_i = \phi_v(h_i)v_i^{\text{init}} + C \sum_{j \neq i} (x_i - x_j) \phi_x(m_{ij}) \]
      \item Position Update: 
        \[ x_i' = x_i + v_i \]
        
\end{itemize}

\section{Experimental Details}
\label{app:details}

\subsection{N-Body System}
\label{app:nbody}

We used the same setup as in \citet{satorras2021n}, i.e. 3,000 training trajectories, 2,000 validation trajectories, and 2,000 test trajectories. Each trajectory contains 1,000 time steps. We used a 4-layer EMPCN of dimension 2 for our experiments, setting the initial number of hidden features to 66, resulting in a total of 200k parameters in total. Additionally, a dropout layer with a probability of 0.2 was used for every single update and message passing. The invariant features are embedded using Gaussian Fourier features as introduced in \citet{tancik2020fourier}. The optimization is done using the Adam optimizer, with a constant learning rate of $\eta = 5 \times 10^{-4}$, a batch size of 100, and weight decay of $10^{-12}$. The loss minimized is the Mean Squared Error (MSE) in the predicted position.

Both the upper and lower adjacencies, as well as the boundaries, were considered for the message passing. The different cell communications that are considered are the following: 0-cell (node) to 0-cell (node), 0-cell (node) to 1-cell (edge), 1-cell (edge) to 0-cell (node), 1-cell (edge) to 1-cell (edge), 1-cell (edge) to 2-cell (ring), and 2-cell (ring) to 1-cell (edge).

The invariants used for message passing are the following: 0-cell to 0-cell: node distance; 0-cell to 1-cell (and vice versa): length of edge; 1-cell to 1-cell: length of sending edge, length of receiving edge, distance from the midpoints of sending and receiving edge; 1-cell to 2-cell (and vice versa): ring approximated radius, ring length, edge length.

\subsection{QM9}
\label{app:qm9}

For the QM9 dataset, we utilized the common split of 100,000 molecules for training, 10,000 molecules for testing, and the remaining molecules for validation. All predicted properties were normalized by subtracting the mean of the target values in the training set and then dividing by the mean absolute deviation (MAD) in the training set to stabilize training.

The models were trained for 1000 epochs each, with the final number of parameters set to one million. For the \textit{decoupled EMPCN} model, the parameters were divided 75:25 between the EGNN and the ring-to-node EMPCN components, respectively. To match the one million parameter target, the number of hidden features in the higher-order message passing was reduced to 44, while the number of layers remained equal to 7. For the node-to-node communication, which is handled by the EGNN, only the norm of the distance was used as an invariant. For the ring-to-node communication, the invariants used were the convex hull volume and area, the norm of the distance between the ring midpoint and each node within the ring, and the ring length. Although one could also use the representations of rings in the final readout, we chose to use only the representations of nodes to ensure a fair comparison with the EGNN model.

The models were optimized using the Adam optimizer with an initial learning rate of $5 \times 10^{-2}$ and a Cosine Annealing learning rate scheduler. The Mean Absolute Error (MAE) was used as the loss function for optimization. We used a batch size of 96 molecules and applied a weight decay of $10^{-12}$. Additionally, dropout layers were included in each message and update layer.

\subsection{CMU Motion Capture}
\label{app:cmu}

As discussed in Section \ref{subsec:cmu}, for this task the goal is to forecast the node positions 30 timesteps into the future based on the positions from a randomly chosen frame. We adopted a similar framework to \citet{kipf2017semisupervised, huang2022equivariant,liu2024clifford}, concentrating on the walking motion of a single subject (subject \#35) over 23 trials. 

As there are no rings present we use the extra topological information from connecting elbow joint nodes with shoulder and palm nodes to create edges and triangles. Additionally, we link hip, knee, and heel nodes to form another set of edges and triangles as firstly introduced by \citet{liu2024clifford}. As mentioned in Section \ref{subsec:cmu}, we also introduce rings of length four and five. These rings connect the two elbows with the two knees, the two arms with the two wrists, the two knees with the two feet, and the two knees with the feet and the hip. The specific connections are as follows: [0, 3, 8], [6, 7, 8], [1, 2, 3], [24, 25, 26], [21, 22, 23], [7, 8, 2, 3], [24, 26, 17, 19], [25, 7, 18, 2], and [6, 7, 8, 1, 2, 3].

The models were trained for 5000 epochs each, with the final number of parameters set to 200K. For the \textit{decoupled EMPCN} model, the parameters were divided 80:20 between the EGNN and the ring-to-node EMPCN components, respectively. To achieve the 200K parameter target, the number of hidden features in the higher-order message passing was reduced to 33 in comparison to 66 that was for the node to node communication, while the number of layers was set to 4. Similar to the experimental details for the QM9 dataset mentioned in Section \ref{subsec:qm9} and Appendix \ref{app:qm9}, for the final readout we used only the representations of nodes.

The models were optimized using the Adam optimizer with an initial learning rate of $5 \times 10^{-2}$ and a Cosine Annealing learning rate scheduler. The loss minimized was the Mean Squared Error (MSE) in the predicted position. A batch size of 100 was used, along with a weight decay of $10^{-8}$. Additionally, dropout layers with a probability of 15\% were included in each message and update layer. Finally, the invariants used for the ring-to-node communication were the norm of the distance for node-to-node communication, the ring length, and the norm of the distance from each point to the midpoint of the respective ring.


\end{document}